\documentclass{article}

\usepackage{PRIMEarxiv}

\usepackage[utf8]{inputenc} 
\usepackage[T1]{fontenc}    
\usepackage{hyperref}  
\hypersetup{
    colorlinks=true,
    linkcolor=blue,
    filecolor=blue,      
    urlcolor=blue,
    citecolor=blue,
}     
\usepackage{url}            
\usepackage{booktabs}       
\usepackage{amsfonts} 
\usepackage{amsmath}
\usepackage{nccmath}
\usepackage{tabularx}
\usepackage{graphicx}
\usepackage{caption}  
\DeclareCaptionFormat{sanslabel}{#3}    
\usepackage{nicefrac}       
\usepackage{microtype}      
\usepackage{lipsum}
\usepackage{fancyhdr}       
\usepackage{graphicx}       
\graphicspath{{media/}}     
\usepackage[symbol]{footmisc}

\title{Toward High-Performance Energy and Power Battery Cells with Machine Learning-based Optimization of Electrode Manufacturing}

\author{ {\hspace{1mm}Marc Duquesnoy}$^{1, 2, }$\footnote[2]{text}\\
	\texttt{} \\
	\And
	{\hspace{1mm}Chaoyue Liu}$^{1, }$\footnote[2]{text} \\
	\texttt{} \\
	\And
	{\hspace{1mm}Vishank Kumar}$^{3}$ \\
	\texttt{} \\
	\And
	{\hspace{1mm}Elixabete Ayerbe}$^{2, 4}$ \\
	\texttt{} \\
	\And
	{\hspace{1mm}Alejandro A. Franco}$^{1, 2, 5, 6, *}$}

\begin{document}
\maketitle

{$^{1}$ \small Laboratoire de Reactivit\'{e} et Chimie des Solides (LRCS), UMR CNRS 7314, Universit\'{e} de Picardie Jules Verne, Hub de l\'{}Energie, 15 rue Baudelocque, Amiens Cedex, 80039, France}

{$^{2}$ \small ALISTORE-European Research Institute, FR CNRS 3104, Hub de l'Energie, 15 rue Baudelocque, Amiens Cedex, 80039, France}

{$^{3}$ \small Umicore, Corporate Research \& Development, Watertorenstraat 33, Olen, 2250, Belgium}

{$^{4}$ \small CIDETEC, Basque Research and Technology Alliance (BRTA), P$^\circ$ Miram\'{o}n 196, Donostia-San Sebastian, 20014, Spain}

{$^{5}$ \small Institut Universitaire de France, 103 Boulevard Saint Michel, Paris, 75005, France}

{$^{6}$ \small Réseau sur le Stockage Electrochimique de l\'{}Energie (RS2E), FR CNRS 3459, Hub de l\'{}Energie, 15 rue Baudelocque, Amiens Cedex, 80039, France}

{$^{*}$ \small Corresponding author : alejandro.franco@u-picardie.fr}

\footnotetext[2]{These authors contributed equally to this work.}

\vspace{2cm}

\begin{abstract}
The optimization of the electrode manufacturing process is important for upscaling the application of Lithium Ion Batteries (LIBs) to cater for growing energy demand. In particular, LIB manufacturing is very important to be optimized because it determines the practical performance of the cells when the latter are being used in applications such as electric vehicles. In this study, we tackled the issue of high-performance electrodes for desired battery application conditions by proposing a powerful data-driven approach supported by a deterministic machine learning (ML)-assisted pipeline for bi-objective optimization of the electrochemical performance. This ML pipeline allows the inverse design of the process parameters to adopt in order to manufacture electrodes for energy or power applications. The latter work is an analogy to our previous work that supported the optimization of the electrode microstructures for kinetic, ionic, and electronic transport properties improvement. An electrochemical pseudo-two-dimensional model is fed with the electrode properties characterizing the electrode microstructures generated by manufacturing simulations and used to simulate the electrochemical performances. Secondly, the resulting dataset was used to train a deterministic ML model to implement fast bi-objective optimizations to identify optimal electrodes. Our results suggested a high amount of active material, combined with intermediate values of solid content in the slurry and calendering degree, to achieve the optimal electrodes.
\end{abstract}


\section{Introduction}
\paragraph{} In our modern society, the demand for batteries has surged due to the widespread use of electric vehicles and portable electronic devices. Lithium-ion batteries (LIBs) have emerged as the most powerful technology for a fast energy transition \cite{1,2}. Driven by the increasing demand for high-performance energy solutions with low-carbon emissions, the modern world is making efforts to establish gigafactories and recycling solutions to significantly reduce the production costs for LIBs and make it sustainable \cite{3,4}. The manufacturing process is considered the most impactful part of battery design, and optimizing this process is crucial for improving overall battery performance \cite{5}. This complex fabrication process involves numerous interlinked steps and manufacturing parameters, resulting in various production approaches \cite{6}. The entire process includes electrode slurry preparation, coating and drying, calendering, and the the cell assembly, electrolyte filling and formation \cite{7}. Certain parameters, such as the type of material, the amount of material, and the drying temperature applied to the slurry, have a significant impact on the final battery performance and must be optimized throughout the entire fabrication process. The electrode optimization in turn depends on the end application that can be categorized as: 1) energy-oriented and 2) power-oriented batteries \cite{8,9}, which require different electrode design strategies. In general, energy-oriented batteries favor higher material loading, while power-oriented batteries favor lower material loading due to the nonlinear relation between electrode energy density and applied current \cite{10,11,12,13}.

\paragraph{} The digitalization of the manufacturing process for LIBs has already demonstrated progress in providing meaningful results to better understand the impact of the manufacturing process chain on electrode properties, such as conductivity, mass loading, and electrode porosity \cite{14,15,16,18,19,20}. This digitalization is supported by physics-based models and Machine Learning (ML) models, which enable high throughput modeling and accelerate the discovery of interlinked manufacturing steps \cite{17, 17_ters}. The ERC-funded ARTISTIC project has become a pioneer in computational 3D-validated models and has developed models for each step in the manufacturing process, with the models sequentially linked with one another \cite{21}. For example, the output from the slurry model is used as the input for the drying model, which is followed by the calendering process simulation. Coarse-Grained Molecular Dynamics (CGMD) and the Discrete Element Method are used for the simulation of the slurry, its drying and the calendering of the resulting electrode \cite{22}. At the electrochemical level, a 4D-resolved continuum model was used to perform electrochemical and mechanical simulations of galvanostatic discharge-charge and electrochemical impedance spectroscopy experiments \cite{23}, allowing us to determine the relationship between performance and the corresponding manufacturing parameters. Additionally, all of these models have been experimentally validated using our battery prototyping platform, and their practical application is enabled through an online calculator service accessible to any user with an internet connection \cite{24}. New applications of ML techniques were also integrated in the ARTISTIC computational workflow, supported on both classification and regression models \cite{17_bis, 17_ters}.\\

\begin{figure}[!ht] 
	\captionsetup{format=sanslabel}
    \hbox to\hsize{\hss\includegraphics[scale=0.33]{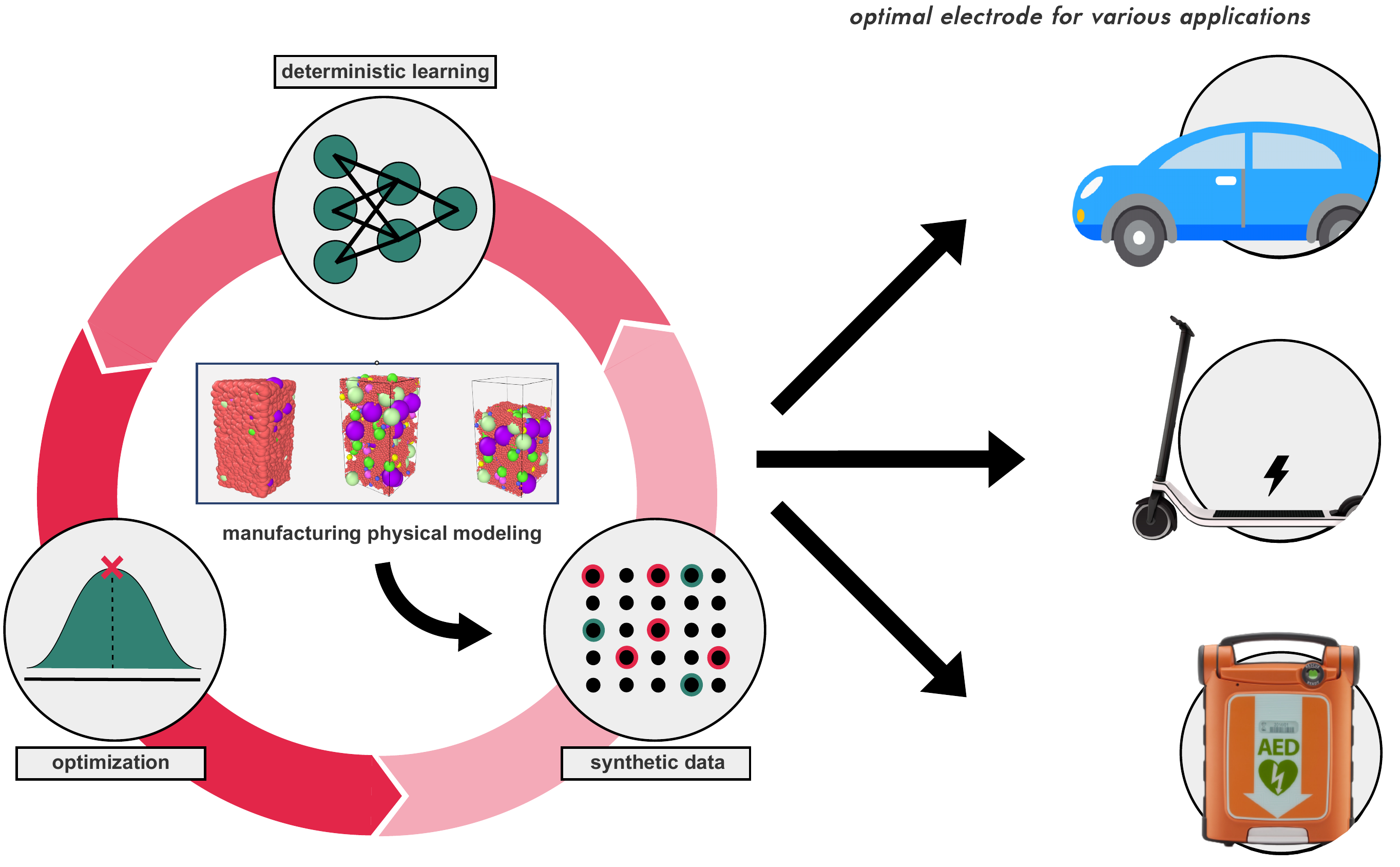}\hss} 
    \caption{\textbf{Figure 1}: Schematic representation of the deterministic approach for our electrochemical performance optimization based on the energy density and power density for different battery applications. The workflow is similar to the one used in Ref. \cite{13} but here it takes into account the optimization of cell performances instead of electrode textural properties.}
\end{figure}

\paragraph{} In general, CGMD simulations can generate an electrode slurry model in as little as eight hours, but NEMD simulations can simulate the viscoelastic behavior of the slurry structure in as little as five days. Due to this restriction, physical modeling cannot be used for real-time optimization of manufacturing processes \cite{25,26}. This limits the research on automatic optimizations of other performance metrics in the interim \cite{27}. Still, physics-based simulations are very important, because they provide, in a high fidelity fashion, an understanding based on physics of the reasons of why and how manufacturing pararemeters impact electrode properties. Moreover, ML algorithms have shown impressive potential in studying the effects of manufacturing processes on electrode properties, such as predicting final electrode properties using regression models \cite{28}, or estimating battery state-of-health using common regression techniques \cite{29}. These algorithms can also support physics-based models in significantly reducing their overall computational costs. For example, Functional Data Analysis has been applied by us to viscosity simulations to reduce the time required for retrieving simulation results \cite{26}. Gao \textit{et al.} developed a hybrid solution combining modeling and data-driven approaches to accelerate electrode design \cite{30}, while Thelen \textit{et al.} integrated physical modeling and ML methods for primary degradation models and battery cell capacity estimation \cite{31}. Furthermore, inverse design of battery manufacturing is a challenging yet promising approach. Lv \textit{et al.} focused on response surface methodology to optimize titanium and vanadium that is co-doped in LiFePO$_4$/C \cite{31_bis,31_ters}. Bayesian Optimization (BO) has been used to solve complex optimization problems with a probabilistic approach \cite{32}, and has been widely applied in domains such as biology for sequence design \cite{33}, and cybersecurity for cyber-attack detection \cite{34}. However, R\&D facilities are required to build even more appropriate devices due to the rising demand for battery development. In that regard, BO is in the spotlight for solving issues of this nature. It is worth noting that maximizing cell energy is limited by high power usage, illustrating the need for multi-objective optimization \cite{35,36}. Similarly, increasing electrode thickness for energy cells can result in faster capacity fade and an expansion of the tortuosity factor \cite{37}. Consequently, the growing interest in inverse design requires further exploration to provide numerical frameworks for various optimization objectives.

\paragraph{} This study extends our previous work on inverse design of manufacturing parameters to new energy- or power-like applications, and proposes a data-driven approach to maximize electrochemical performance. To generate a synthetic dataset of 3D calendered microstructures of LiNi$_{1/3}$Mn$_{1/3}$Co$_{1/3}$O$_2$ (NMC111) active material-based electrodes, we employed the chain of ARTISTIC physics-based models that link input manufacturing conditions and output electrode properties, as reported in our previous work \cite{13}. These properties were treated as input parameters of a Newman's pseudo-two-dimensional (p2D) model to produce electrochemical performances under galvanostatic discharge. The obtained results offer an appropriate dataset to train and test deterministic learning models, with primary objectives aimed at approximating the complete computationally expensive electrode physics-based generation and characterization and enabling interpolating the manufacturing inputs space in order to analyze other conditions that were not included in the initial synthetic experiments. By blending these functions into a BO framework for bi-objective optimizations, we identified the best amount of active material (AM\,\%), slurry solid content (SC\,\%), and calendering compression degree (CD\,\%) that can give an electrode with a relevant energy and power densities for various cell applications as it can be illustrated in Figure 1. The bi-objective optimizations were selected based on the specification of energy-like or power-like applications at the electrode level that were mathematically expressed for such NMC111 material, similar to concrete uses. These potential applications are now being researched and developed, which aims to find ways to optimize cell performance for a range of issues at the lab and industrial levels. That is why our approach focuses on the comparison of the bi-objectives to facilitate the design of battery problems and to emphasize the importance of assessing an inverse design loop. This approach opens up the possibility of developing autonomous battery manufacturing processes.\\

\section{Manufacturing physics-based modeling}
\subsection*{Data acquisition}
\paragraph{} Due to the high computational cost required to simulate the electrochemical performance for a continuum batch of various manufacturing conditions, we utilized the synthetic dataset generated from our previous work \cite{13} to obtain a highly representative dataset of the manufacturing parameter space. Specifically, we have generated quasi-random Sobol sequences with Saltelli extension based on three parameters: the amount of active material (AM\,\%), the slurry solid content (SC\,\%), and the electrode compression degree (CD\,\%) \cite{39}. These parameters are representative enough of the slurry preparation, drying, and calendering processes, as important parameters to assess when manufacturing electrodes \cite{39_bis,39_ters}. Our focus was to properly probe the input manufacturing space and capture all of its sub-areas by varying these three parameters. This design of experiments (DOE) was used as input values for physics-based models to evaluate numerous electrode properties that characterize the 3D microstructures \cite{40}. More details on this can be found in section 2 of our previous work \cite{13}. It is worth mentioning that the DOE is large enough to generate data for further machine learning (ML) regression purposes while being efficient enough to avoid a huge computational cost associated with generating all 3D microstructures for each manufacturing condition. In fact, using formulation and drying as input parameters, the production of a slurry and its matching dried electrode can take in computing clusters roughly 150 hours. When using DOE approaches to map the behavior of the electrodes under manufacturing circumstances as precisely as feasible, the challenge of producing huge batches of synthetic data may be overcome. In the end, the DOE ultimately includes 174 alternative situations that were deemed relevant for the analysis of the manufacturing input space with the chosen parameters in our earlier work, which was centered on enhancing electrode textural properties.\\

\begin{figure}[!ht] 
	\captionsetup{format=sanslabel}
    \hbox to\hsize{\hss\includegraphics[scale=0.35]{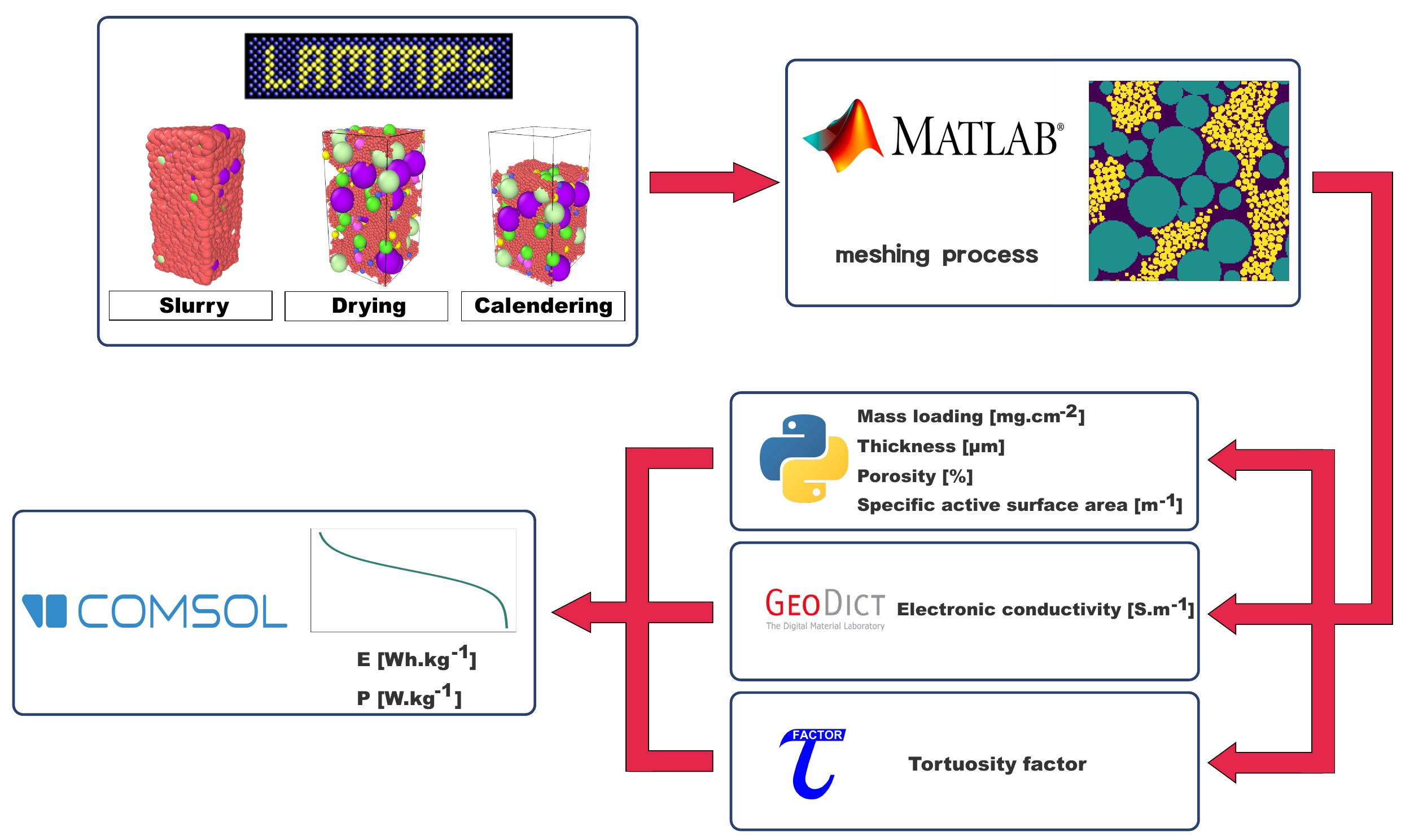}\hss} 
    \caption{\textbf{Figure 2}: Schematic representation of the link between the different softwares for the handling of the 3D electrode microstructures for the data acquisition generated by the physics-based modeling workflow. This starts with the simulations from the slurry to the calendering, where the resulting electrode is meshed in order to serve as input of the different codes for the properties calculation. This ends by using the latter properties as inputs of a p2D model to simulate the electrochemical performances to then extract the gravimetric energy density and and the average gravimetric power density.}
\end{figure}

\paragraph{} In this work, we went beyond the study we previously reported by using the electrode properties calculated from the 3D calendered microstructures as input parameters of a Newman's pseudo-two-dimensional (p2D) model in order to simulate the electrochemical behavior of the electrode at 1C discharge like Figure 2 illustrates. Six properties characterized from the synthetic electrode microstructures were fed into the  p2D model: electrode porosity, tortuosity factor, mass loading, specific active surface area, electronic conductivity, and thickness. In our electrochemistry simulation, the CBD phase is considered to block Li$^+$ transport. Therefore, the nanoporosity of CBD is not taken into account for electrode porosity calculations and specific active surface areas only consider the interface between AM and electrolyte. Both values are calculated based on the voxelized microstructures (Figure 2). Tortuosity factor and electronic conductivity are acquired by the open-source code TauFactor \cite{41} and the commercial code GeoDict respectively \cite{42}. Since the p2D model is a widely used model for LIBs study, we do not discuss much more on details of the associated p2D modeling setup. The equations and parameters used are listed in Table S1 and Table S2 in the Supporting Information. Worth to be noted that the size distribution of AM particles is taken into consideration in our CGMD model and p2D model, which is presented as 6 groups of particles with different sizes. All cases have the same particle size distribution profile. Volume fractions are reported in Table S2 in the Supporting Information individually. The specific active surface area is also divided into six groups by the equation:\\

\begin{ceqn}
\begin{align}
a_{v,n} = \frac{r_n^2n_n}{\sum_{m} r_m^2n_m} a_v\epsilon
\end{align}
\end{ceqn}

\noindent where $a_v$ is the total specific active area acquired from voxelized structures, $\epsilon$ is the volume fraction, $r_m$ is the radius of the particle and $n_m$ is the number of particles of each type.\\

\begin{figure}[h] 
    \captionsetup{format=sanslabel}
    \hbox to\hsize{\hss\includegraphics[scale=0.45]{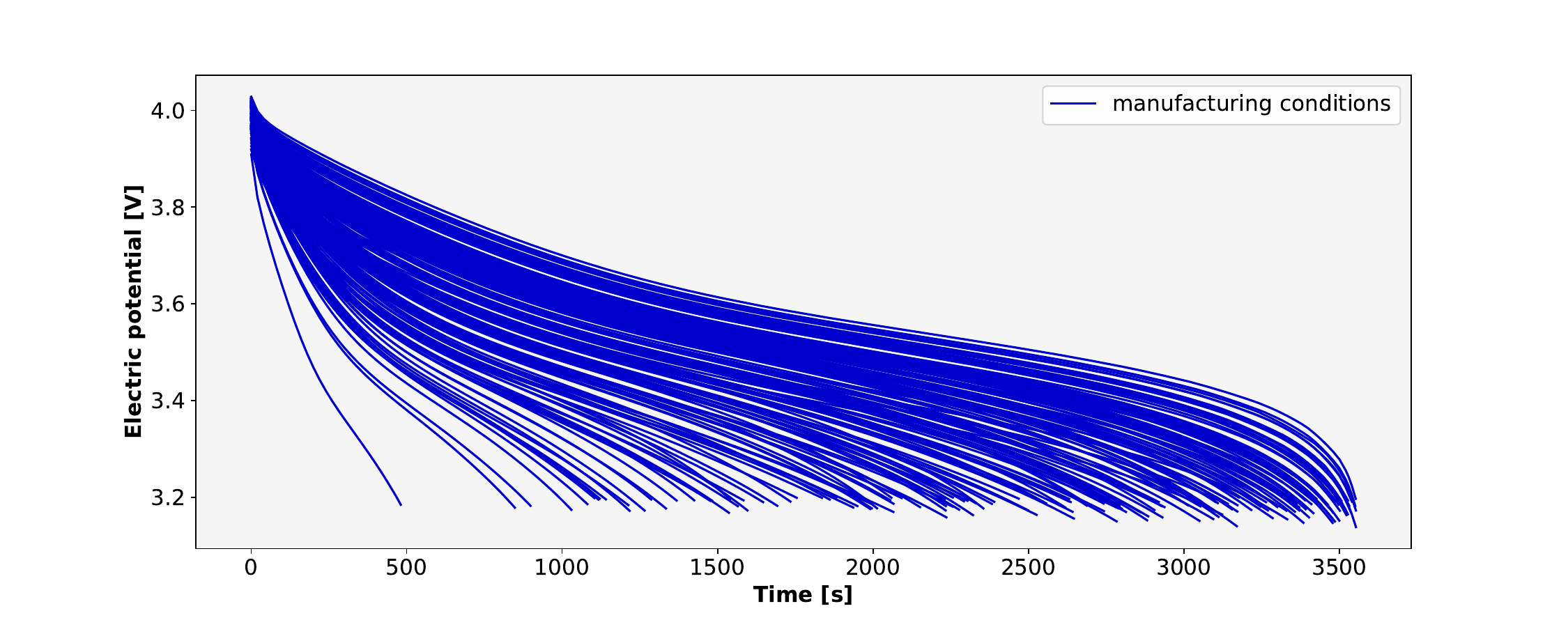}\hss}
    \caption{\textbf{Figure 3}: 1C discharge curves resulting from the simulation of the electrochemical performances using the Newman's p2D model in COMSOL Multiphysics 5.6 for the different electrode microstructures generated by the ARTISTIC manufacturing simulator.}
\end{figure}

\paragraph{} Simulations of 1C discharge with a cut-off voltage of 3.2\,V were performed for all microstructures through parameter sweep. The discharge curves are displayed in Figure 3. The difference in voltage plateaus and discharge capacities shows the effect of microstructural characteristics on the electrochemical performance. What leads to this difference is the overpotential. The source of overpotential could be originated from four aspects: mass transport in the solid phase, mass transport in the liquid phase, ionic, electronic resistance, and Faradaic reaction overpotential \cite{42_bis}. Since we have the same particle size distribution for all cases, the solid phase mass transport is under the same condition and therefore it is not the source of differences. Mass transport in the liquid phase, however, is the main source of capacity decrease. Figure S1 in the Supporting Information shows the distributions of properties of all cases. The variety of porosities and tortuosity factors shown in Figure S1 causes considerable differences in Li$^+$ concentration distribution in the electrode. Low concentration leads to low exchange current density in the Butler-Volmer equation, which returns a high Faradaic overpotential locally. Thus, the cases got lower discharge capacities as the Li$^+$ concentration gradient went higher.

\paragraph{} Electronic resistance was found to be not a determining factor in capacity difference. As we have reported in our group's previous work, an obvious indicator can be used to identify if electronic conductivity is the limiting factor \cite{23}: if the particle SOC is found to be higher at the current collector side, it means that the electrode has a low electronic conductivity which causes capacity decrease. However, this phenomenon was not found in our study, meaning that electronic resistance is not the main reason leading to the capacity decrease. The specific reaction area is different due to the different AM formulations in the synthetic electrodes. A higher AM ratio means a more exposed AM surface to electrolyte due to less coverage of CBD. This will lead to lower local surface Faradaic current, therefore lower reaction overpotential.

\paragraph{} We recognize that the discharge profile of an electrode reflects a complex combination of characteristics. Discharge capacity is insufficient to comprehensively demonstrate the electrochemical performance, thus we use two other descriptors, gravimetric energy density $E$ and average gravimetric power density $P$, to measure and compare electrode performances. Conventionally, $E$ and $P$ are referred in the community for the full battery cell to describe the energy and power oriented cell design, but here we are referring them in respect to a single electrode only as descriptors for electrode performance assessment. Readers must avoid conflating the two concepts.\\

\begin{ceqn}
\begin{align}
E = \frac{\int I_{app}Vdt}{M_{electrode}}
\end{align}
\end{ceqn}
\begin{ceqn}
\begin{align}
P = \frac{E}{t_{total}}
\end{align}
\end{ceqn}
 
\noindent where $I_{app}$ is the applied current, $V$ is the output voltage, $M_{electrode}$ is the total mass of the electrode including mass of AM, CBD, electrolyte and current collector. $E$ and $P$ will be the optimization targets of our BO framework. The gravimetric energy density $E$ could be used individually to determine the optimum case. By adding $P$ as another analysis dimension, more information could be extracted enabling extending the optimization to other C-rates. As we know, the gravimetric energy density $E$ can by itself determine the optimum electrode for a given current condition, which in our case is 1C discharge. However, $E$ alone is not enough to capture the characteristics of electrode under other conditions without doing electrochemical simulations. Combining with $P$, the cases can be separated into two groups that correspond to electrodes that are suitable for higher C-rates and lower C-rates, respectively. In our work, we have an optimization problem that find the optimal electrodes at 1C whereas two other scenarios focus on an electrode that should be operated at low and ultra-low C-rates. More detail discussion will be given in later sections.

\subsection*{Deterministic learning}
\paragraph{} The resulting synthetic dataset gathered enough manufacturing parameters and associated electrode properties to train and validate deterministic learning, serving as numerical functions to calculate the electrode performances at the electrochemical level without the need to launch any new physics-based modeling simulation. Indeed, such learning constitutes regression functions trained on pre-defined data using the aforementioned DOE, improving the predictive capabilities to forecast manufacturing conditions from the input parameter space \cite{38}. Therefore, the time needed to retrieve the electrode performances is drastically reduced because it takes only a few seconds compared to the usual case within the full physics-based model chain. This possibility to inter/extrapolate for any relevant manufacturing condition which is not represented within the DOE enables us to carefully initialize the inverse design and save time when analyzing extra conditions. In that sense, the combination of the DOE and predictive models capture the sensitivity of the relationships between the selected properties and the manufacturing parameters which is meaningful to optimize multiple properties simultaneously \cite{42_ters}. We used a Gaussian process (GP) applied for non-linear regression purposes to fit the latter two properties as a function of the manufacturing parameters, due to its potential to infer a probability distribution over the training dataset as a collection of random variables any finite number of which have a joint Gaussian distribution \cite{43,44}. It is particularly efficient for us since the training dataset is highly representative of the entire manufacturing parameter space to estimate the prior distribution. The associated algorithm assimilates any output property $y$ (\textit{i.e.}, the energy or power density) written as $y = f(x) + \epsilon$ where $x$ is a set of input values (\textit{i.e.}, a manufacturing condition) and $\epsilon$ a noise term with variance $\sigma_{\epsilon}^2$.\\

\paragraph{} However, $f$ is an unknown function which will be here approximated, assuming a Gaussian process 

\begin{ceqn}
\begin{align}
f(x) \sim \mathcal{GP}(m(x), k(x, x^{'}))
\end{align}
\end{ceqn}

\noindent where $\mathcal{GP}$ is defined by a mean $m(x)$ and a covariance function $k(x, x^{'})$ which depends on a \textit{kernel} function written $k$ \cite{45}. The choice of the kernel function depends on the relationships between input variables and the smoothness that exists in the patterns of data. It is common to use the \textit{radial basis function} (RBF) \cite{46}, the \textit{constant kernel} (CK), the \textit{exponential sine squared kernel} (ESSK) \cite{47}, or even a combination of any existing kernels. For our case study, we choose the RBF whose free parameter is equal to 1 because $k(x, x^{'})$ has to decrease smoothly with distance for sparse inputs, whereas having similar outputs for similar inputs.

\paragraph{} Under prior distribution from the training dataset $(x_i, y_i)_{\{i \leq n\}}$ (where $n$ represents the number of training data points), $y = [y_1, y_2, y_3, ..., y_n]_{\{i \leq n\}}$, and the kernel $k$, GP assigns the posterior distribution by evaluating a joint Gaussian distribution for any new testing data point $x^*$ as it follows matrix-wise spoken \cite{48,43}\\

\begin{ceqn}
\begin{align}
\left[ {\begin{array}{c}
    y \\
    y^* \\
  \end{array} } \right] \sim \mathcal{N}(0, \left[ {\begin{array}{cc}
    k(x,x) + \sigma_{\epsilon}^2 I_n & k(x,x^*) \\
    k(x,x^*) & k(x^*, x^*)\\
  \end{array} } \right])
\end{align}
\end{ceqn}

\noindent where $k(x,x)$ is the $n$-order covariance based on the training set, $k(x,x^*)$ is the covariance based on the training set and the new testing data point $x^*$, and $y$ is the set of output properties defined above.\\

\paragraph{} Therefore, it becomes feasible to predict the output corresponding to the new testing data point $x^*$ according to the posterior probability distribution which is also following a Gaussian distribution but with different mean and covariance functions as written below:

\begin{ceqn}
\begin{align}
    \begin{cases}
    y^* | x^* \sim \mathcal{N}(\mu, \sigma) \\
    \mu = k(x, x^*)[k(x,x) + \sigma_{\epsilon}^2 I_n]^{-1} y \\
    \sigma = k(x^*,x^*) - k(x^*,x)[k(x,x) + \sigma_{\epsilon}^2 I_n]^{-1}k(x,x^*)
    \end{cases}
\end{align}
\end{ceqn}

\paragraph{} That is being said, the new output can be rewritten as a linear sum of similarities between the new testing data point and each training data point from the kernel application, weighted by coefficients $w_i = [k(x_i,x_i) + \sigma_{\epsilon}^2 I_n]^{-1} y_i$ that arise from the numerical calculations of covariances matrix from equation (6) \cite{48}. It results a new equation for the mean of the posterior distribution $\mu = \sum_{i=1}^{n} w_i \times k(x_i, x^*)$.

\subsection*{Optimization process}
\paragraph{} After training the deterministic learning for electrochemical performance prediction, we performed Bayesian bi-objective optimization loops (BO) by assessing a specific objective function (\textit{i.e.}, $C_f$)  being dependent on the latter numerical functions \cite{49,50}. BO intended to optimize the energy and power density differently by formalizing the objective function into a single-weighted minimization problem according to the final battery application. BO is already well known for multi-objective optimizations, applying Gaussian process regressions as the model to numerically approximate $C_f$ \cite{51}. BO represents an iterative loop over data that have been tested as potential candidates of the optimal conditions (\textit{i.e.}, $\mathcal{D}$), where the algorithm calculates a posterior distribution ($C_f$) | $\mathcal{D}$ to propose a new set of data (\textit{e.g.}, a set of manufacturing parameters) through an acquisition function which balances between the exploitation of prior parameter value combinations from $\mathcal{D}$ to establish nearby minima, and the exploration to identify minima far from prior parameter value combinations. In the end, the optimization loop stops when $C_f$ has reached values under a certain threshold, or after a particular number of steps that have been pre-defined when initializing the BO algorithm. Considering that optimizing battery manufacturing at the electrochemical level means maximizing the energy and power density for our study, we formalized a minimization problem with $C_f$ which is equivalent to find the best manufacturing parameter combination $\mathbf{x}^*$ that minimizes the objective function expressed like \\

\begin{ceqn}
\begin{align}
  \mathbf{x}^* \hspace{0.1cm} = \hspace{0.1cm} argmin_x \hspace{0.09cm}(C_f(x))
\end{align}
\end{ceqn}

\subsection*{Relative importance assessment}
\paragraph{} For every bi-objective optimization performed in this study, both $E$ and $P$ are considered simultaneously. However, we granted different relative importance (called \textit{weights} in the following paragraphs) according to the problem we aimed to solve. To obtain the best electrode, in other words, the electrode with the highest energy density, we put full weight on $E$ and zero on $P$. Then, more weight is assigned to $P$ to shift the optimum case to the high $P$ direction. As a result, several sets of manufacturing parameters will be acquired. The following equation (8) denotes how $C_f$ was built to correspond to the different weighted power bi-objective optimizations \cite{54}.\\

\begin{ceqn}
\begin{align}
  C_f(x_i) = w_{E} \times (1 - y_{\{x_i, E\}})^2 + w_{P} \times (1 - y_{\{x_i, P\}})^2
\end{align}
\end{ceqn}

\paragraph{} Indeed in the BO loops, the deterministic learning predicts the $E$ and $P$ for a given set of manufacturing parameter values $x_i$ at each step, which is quick and reduces the computational cost for any optimization. Then, we applied a scalar fitness function to transpose the numerical predictions from the deterministic learning (\textit{i.e.}, $y_{\{x_i, E\}}$ and $y_{\{x_i, P\}}$) into the interval [0, 1] when calculating the two terms of the $C_f(x_i)$ for a given set of manufacturing parameter values $x_i$ (the scalar fitness function is reported in the Supporting Information). Overall, this gives relative order of magnitude for standardized terms and then avoids a bias induces by the values of $E$ and $P$. In the end, the maximization of the two electrode performances is equivalent to minimize the terms $1 - y_{\{x_i, E\}}$ and $1 - y_{\{x_i, P\}}$ to target the lowest $C_f$ as possible. The weight of each term (\textit{i.e.}, $w_{E}$ or $w_{P}$) will be given by the type of battery-oriented problem (Table 1). Two extra cases are given to show the trend of electrode performance change with weight assignment. All these problems correspond to a tremendous practical value for labs and industries. They have been chosen for a comparative purpose where other possible battery-oriented problems can be formalized using different weights with respect to real applications.

\paragraph{} Nevertheless, the assignment of weights must reflect accurate quantification regarding the decision-making, which is difficult to evaluate in reality since they may depend on the overall prediction error \cite{55}. That is why it exists a variety of proposals to determine the correct weights in a multi-criteria approach \cite{56}. One popular way is to give equal importance to all of the properties to be optimized in the BO loop when the decision maker knows nothing about the right weights, whereas unbalanced and ranked approaches seem to be more convenient when the order of importance is at least prioritized \cite{57}. For instance, the \textit{rank sum weight method} (RS), the \textit{rank exponent weight method} (RE), or the \textit{rank-order centroid weight method} (ROC) are relevant alternatives once the ranked attribute weights are pre-defined \cite{57}.\\

\begin{table}[h]
\centering
\small
\setlength\tabcolsep{3pt}
\setlength\extrarowheight{2pt}
\captionsetup{format=sanslabel}

\begin{tabularx}{\textwidth}{ 
  >{\raggedright\arraybackslash}X 
  >{\raggedright\arraybackslash}X 
  >{\raggedright\arraybackslash}X}
\midrule
\textsc{optimization problem} & \textsc{Method} & \textsc{Weights}       \\ \midrule
optimal electrode at 1C & RE & \{1, 0\}\\
    electrode should be operated at low C-rate & equal & \{$\frac{1}{2}$, $\frac{1}{2}$\}\\
    electrode should be operated at ultra low C-rate & RE & \{0, 1\}\\
    \midrule
    case ($*$) & RE & \{$\frac{7}{8}$, $\frac{1}{8}$\}\\
    case ($**$) & RE & \{$\frac{1}{8}$, $\frac{7}{8}$\}\\
    \midrule          
\end{tabularx}
\caption{\textbf{Table 1} : Assessment of weights for the inverse design based on the type of optimization problem. It is important to notice that the sum of weights must be equal to 1. The \textit{RE} weights method corresponds to the \textit{rank exponent weight method} to assess the weights for the bi-objective optimizations. When the relative importance of both properties must be the same, we said \textit{equal} weights.}
\end{table} 

\paragraph{} Since we have two performance descriptors in our study, it is obvious to define the order of the relative importance for the objective function assessment. We selected the RE approach to assign the two weights (equation (9)) as a generalization of the rank sum method since we decided to have a steeper distribution of weights in order to correctly define the battery-oriented optimizations with larger differences among weights \cite{56}. Last but not least, it is straightforward that extreme weight differences induce a bias in the bi-objective (and in a multi-objective in general) optimization, being similar to a single-objective optimization. Therefore, the BO can be insensitive to the distribution of weights, predicting manufacturing parameter conditions. As an example in our case study, the optimal electrode at 1C was found with a single objective optimization similar to an extreme bi-objective optimization on $E$.
\\

\begin{ceqn}
\begin{align}
  w_i = \frac{(2-i+1)^p}{\sum_{k=1}^{2} (2-k+1)^p}  \quad p \in \mathbb{N}
\end{align}
\end{ceqn}

\subsection*{Validation of the deterministic learning}
\paragraph{} The deterministic learning used the DOE which was split into a training and testing dataset for regression training and validation respectively. 80\,\% of the total raw data randomly selected were affected by the training dataset, while the remaining 20\,\% went to the testing dataset. The ratio 80\,\%/20\,\% was empirically chosen but is very common as a standard practice in ML \cite{58}. The R$^2_{score}$ and the root mean square error in percentage (RMSE\,\%) were chosen as the validation metrics in order to analyze the goodness of fit for both types of datasets. Table 2 reports the validation metrics for each electrode performance included in the deterministic learning, where Figure 4 supports accurate results from the overall table. Indeed, an R$^2_{score}$ close to 1 highlights good capabilities for the deterministic learning to predict the energy and power density whereas the RMSE\,\% explicit how far a prediction is from its real synthetic value in average (expressed percentage).\\

\newpage
\begin{table}[h]
\centering
\small
\setlength\tabcolsep{3pt}
\setlength\extrarowheight{2pt}
\captionsetup{format=sanslabel}

\begin{tabularx}{\textwidth}{ 
  >{\raggedright\arraybackslash}X 
  >{\raggedright\arraybackslash}X 
  >{\raggedright\arraybackslash}X 
  >{\raggedright\arraybackslash}X}
\midrule
\textsc{Property} & \textsc{RMSE\,\%} & \textsc{ R$^2_{score}$} & \textsc{CI95}       \\ \midrule
    Training process& & & \\
	E & 4.986 & 0.980 & [0.993; 0.995] \\
	P & 0.968 & 0.986 & [0.991; 0.993] \\
	\midrule    
	Testing process & & & \\
	E & 3.490 & 0.975 & [0.982; 0.985] \\
	P & 1.002 & 0.984 & [0.983; 0.986] \\
    \midrule          
\end{tabularx}
\caption{\textbf{Table 2}: Validation metrics calculated over the training and testing dataset, associated with the fitting of the energy performance $E$ and power performance $P$. The 95\,\% confidence interval (CI95) were estimated with a total of 75 random seeds of training/testing datasets for the uncertainty on the R$^2_{score}$. $E$ and $P$ are coming from equation 2 and equation 3 and correspond to the gravimetric energy density and average gravimetric power density respectively.}
\end{table}

\paragraph{} Even though the deterministic learning is obtained after training and testing the GP regression models, we repeated such a process 75 times by shuffling uniformly the training and testing dataset to calculate the distribution of the R$^2_{score}$ and determine the 95\,\% confidence interval (CI95) \cite{59}. This random selection avoided data leakage and bias when splitting the dataset. This interval contributes to the statistical analysis of how the predictive capabilities of the deterministic learning behave concerning the seed of the re-sampling procedure training/testing dataset. Therefore, CI95 estimates the variability of the metric and how precise it is likely to be given a certain threshold. The larger the CI95 around the mean value, the higher the variability of the deterministic learning to provide accurate predictions regardless of the different training/testing datasets. Lastly, it is usual to conclude that exists a bias in the dataset if the variability is high (\textit{e.g.}, outliers in the training or testing dataset, too few data points, wrong ML model assessment) \cite{60}. Regarding our results, the CI95 includes values close to 1 without a significant width and suggests that deterministic learning can understand the different patterns of data from the training and testing dataset in order to provide accurate predictions.

\subsection*{Optimal manufacturing parameter values}
\paragraph{} The BO framework was set with 300 iterations for each bi-objective optimization as a cut-off to predict the most appropriate candidate for manufacturing parameter combinations for each investigated optimization problem. The function to minimize the posterior distribution within the BO framework was a probabilistic choice based on the lower confidence bound, negative expected improvement, and negative probability of improvement \cite{61}. We initialized the manufacturing conditions generation by a Latin hypercube sequence to start the optimization loop \cite{62}. Table 3 provides the so-predicted best manufacturing conditions associated with the different battery cell applications. On one side, it can be seen that the AM\,\% is always given at a high value, which corresponds to the maximal value that the parameter can take in the BO framework. On the other side, the SC\,\%, and CD\,\% are given at intermediate values compared to the boundaries initialized by the BO framework. These two parameters have been set in the interval [43; 72.8] and [1.4; 38.8] respectively. Therefore, we were capable to obtain the optimal values of the three manufacturing parameters that maximize the energy density under applied C-rate and other C-rates by adjusting relative weights on $E$ and $P$ like Figure 5B compares. Figure S2 and S3 from the Supporting Information support the comparison from Table 3 and Figure 5 to highlight how each predicted manufacturing parameter, $P$, and $E$ are represented within the overall distribution from the synthetic dataset. Even though extreme AM\,\% are always predicted based on the different optimization problems, the latter have a significant impact on the differences in $E$ values rather than $P$ values according to the empirical skewed distribution. That means the bi-objective optimization has to carefully assess the weights in order to avoid a strong decrease of $E$ when optimizing at lower C-rates.

\newpage
\begin{figure}[h] 
    \captionsetup{format=sanslabel}
    \hbox to\hsize{\hss\includegraphics[scale=0.71]{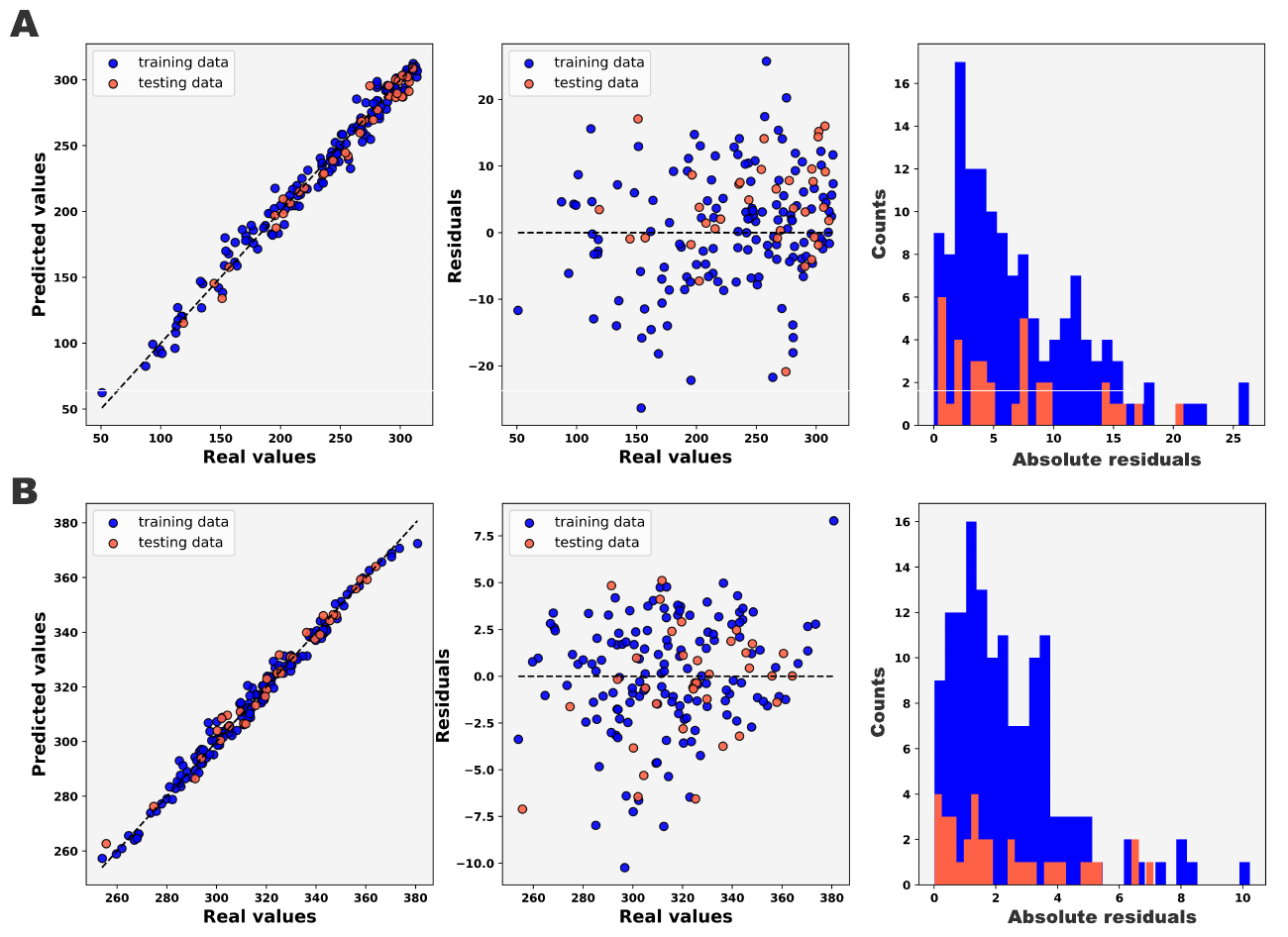}\hss}
    \caption{\textbf{Figure 4}: Illustration of the validation metrics for \textit{(A)} the energy performance and \textit{(B)} the power performance, for the training and testing datasets. For each case, the regression plot displays how values predicted by the deterministic learning are close to real synthetic values. Associated $R^2_{score}$ are reported in the third column of Table 2. The residuals as a function of the real synthetic values show that it does not exist any bias in the training/testing process from the deterministic learning. Lastly, the histogram of absolute residuals displays the distribution differences between predictions and real synthetic values, giving low error predictions on average with some higher errors for a few amount of data points.}
\end{figure}

\section{Results and discussions}
\subsection*{Interest of bi-objectives}
\paragraph{} By introducing $P$ into the analysis, we are able to do multi-C-rate optimization with simulation results with a single C-rate by assigning different relative weight on $E$ and $P$ as in Table 1. As a reminder for the readers, the shift of relative weight should not be connected with the concept of energy-oriented or power-oriented cells. Our optimization work is centered on energy density of electrode, with P as a helper descriptor for further assessments. As we can see in Figure 5A, $E$ and $P$ shows a volcano-shaped outline, where on the peak of the volcano lies the optimal case which has the highest energy density. From the optimal point as the center, the energy density of the electrode decreases monotonically on both sides. With the usage of $P$, the decrease in energy density can be divided into two branches. On the left branch (LB) of Figure 5A, $E$ and $P$ have a positive correlation, while on the right branch (RB) they have a negative correlation. The different $E$ vs. $P$ correlations observed in these two branches have different origins. On LB, the decrease of $E$ comes from the low mass loading of the electrode. Recalling equation (2), the calculation of $E$ takes into account the weight of the current collector and electrolyte. When the electrode mass loading is lower than a threshold, the energy density drops because of the lower mass fraction of active material. Nevertheless, having a low mass loading results in a low diffusion resistance of Li$^{+}$. As a result, the discharge times remain approximately at 3600 s. Therefore, according to equations (2) and (3), $P$ and $E$ form a strong positive linear relation. RB corresponds to high mass loading conditions. The high overpotential causes capacity decrease of the electrode. According to equation (3), decreased discharge time causes an increase of $P$. Therefore, we find a negative relation between $E$ and $P$ on RB. From the analysis above, we can see that mass loading is the key point in the $E$ and $P$ relationship. The conclusion that can be drawn from here is that cases on LB are suitable for higher C-rates, while cases on RB are suitable for lower C-rates. Therefore, in our bi-objective optimization, we categorize the optimization problems as optimal electrode, electrode that should be operated at low-C-rate and ultra low-C-rate objectives, since we are exploring the RB where $P$ increases.\\

\begin{table}[h]
\centering
\small
\setlength\tabcolsep{3pt}
\setlength\extrarowheight{2pt}
\captionsetup{format=sanslabel}

\begin{tabularx}{\textwidth}{ 
  >{\raggedright\arraybackslash}X 
  >{\raggedright\arraybackslash}X 
  >{\raggedright\arraybackslash}X 
  >{\raggedright\arraybackslash}X}
\midrule
\textsc{optimization problem} & \textsc{AM\,\%} & \textsc{SC\,\%} & \textsc{CD\,\%}       \\ \midrule
optimal electrode at 1C & RE & \{1, 0\}\\
    optimal electrode at 1C & 96.8 & 50.8 & 12.8 \\
	electrode should be operated at low C-rate & 96.8 & 60.4 & 16.4 \\
	electrode should be operated at ultra low C-rate & 96.8 & 72.6 & 35.3 \\
	\midrule
	case ($*$) & 96.8 & 56.3 & 13.7 \\
	case ($**$) & 96.8 & 66.4 & 21.8 \\
    \midrule          
\end{tabularx}
\caption{\textbf{Table 3}: Validation metrics calculated over the training and testing dataset, associated with the fitting of the energy performance $E$ and power performance $P$. The 95\,\% confidence interval (CI95) were estimated with a total of 75 random seeds of training/testing datasets for the uncertainty on the R$^2_{score}$. $E$ and $P$ are coming from equation 2 and equation 3 and correspond to the gravimetric energy density and average gravimetric power density respectively.}
\end{table}

\subsection*{Optimized transport properties}
\paragraph{} The optimization results in Table 3 show the best manufacturing parameters we got for each scenario. Going in the direction of lower C-rates results in higher SC\,\% and CD\,\%, while AM\.\% remains at the highest value. Figure S5 from the Supporting Information shows the relation between manufacturing parameters with electrode properties, from which we can see how manufacturing parameters are linked with the electrochemical performances. The main effect of increasing AM\,\% is decreasing thickness and increasing specific surface area. Decreased thickness shortens the diffusion distance of Li$^{+}$ in the electrolyte, and increased specific surface area reduces the local Faradaic current density, which reduces the reaction overpotential \cite{64}. Furthermore, increased AM\,\% reduces the mass fraction of CBD, which increases the calculated $E$ value directly. Thus, it shows a beneficial trend for both energy and power density.

\paragraph{} SC\,\% shows a clearer effect. Because we applied a certain range constraint on coating comma gap, hence the solid content has a linear correlation resulted in mass loading as can be seen in the same Figure S3. In the meantime, electrode thickness also increases with higher solid content. This causes discharge capacity decrease, which is directly linked with the decrease of energy density shown in the plot. However, due to the fact that solid content does not have an obvious effect on other properties, especially specific surface area, the voltage plateau is not affected by solid content. Therefore, based on equation (3), $P$ increases because of the decreased discharge time. Naturally, we got the same trend in the optimized result. 

\paragraph{} The compression degree shows a minor but similar trend as solid content. Figure S3 shows that electrode compression has a complex impact on electrode properties. On one hand, reduced porosity and increased tortuosity will cause poor Li$^{+}$ transport in the electrolyte phase, on the other hand, reduced thickness is beneficial for the Li$^{+}$ transport, and electronic conductivity is also improved due to more compactness of the microstructures and more particle cohesion. The final result depends on the sophisticated competition between the factors. However, our optimization result shows that lower CD\,\% is preferred by the optimal scenario, and higher CD\,\% by the lower C-rate cases. This means that $E$ oriented optimization tends to avoid bad diffusivity of Li$^{+}$ brought by high SC\,\% to make sure full utilization of the active material. $P$ oriented optimization, on the other hand, tends to have a low discharge time in order to promote average power output. Figure 5C exhibits the 3D microstructures at the calendering level generated by the ARTISTIC physics-based manufacturing simulator using the so-predicted best manufacturing conditions from Table 3 as inputs of the physics-based models for each optimization. These 3D volumes consider distinct bottom surface areas and thicknesses due to the differences in terms of formulation and calendering degree.

\newpage 
\begin{figure}[h] 
    \captionsetup{format=sanslabel}
    \hbox to\hsize{\hss\includegraphics[scale=0.085]{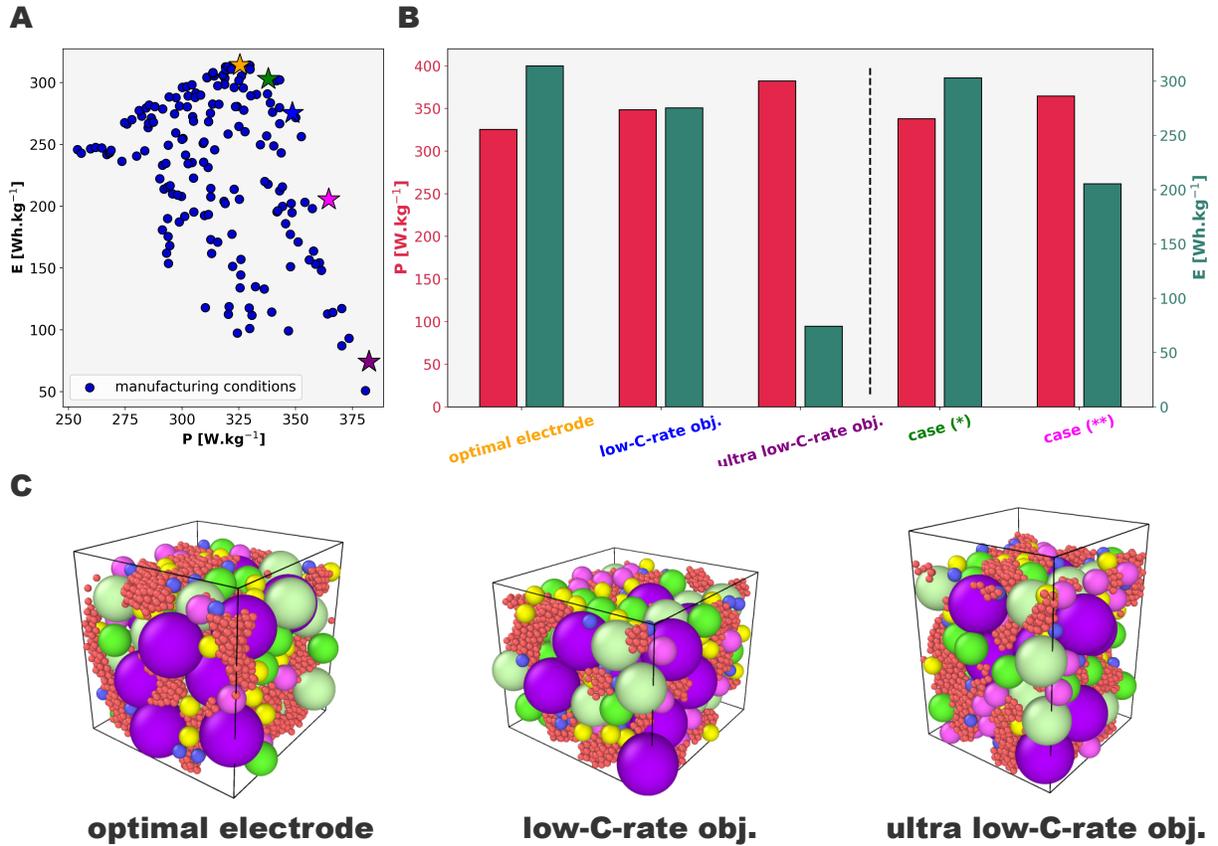}\hss}
    \caption{\textbf{Figure 5}: \textit{(A)}: Scatter plot comparing gravimetric energy density as a function of average gravimetric power density calculated through the p2D model at 1C discharge. The stars illustrate the different optimization problems keeping the same colors as in \textit{(B)} to simplify the legend. \textit{(B)}: Barplot comparing the energy density (color-coded in darkcyan) and the power density (color-coded in red) on the $y$ axis as a function of the type of optimization problem. There are three bi-objective optimizations (\textit{i.e.}, \textit{optimal electrode}, \textit{low-C-rate obj.}, and \textit{ultra low-C-rate obj.} and two extra optimizations to show the trend of $E$ and $P$ as a function of weights. \textit{(C)}: Microstructures of the optimized electrodes. With a purpose of simplifying the graphics, we narrow the two scenarios associated to an electrode that should be manufactured at low and ultra-low C-rates with \textit{low C-rate obj.} and \textit{ultra-low C-rate obj.} respectively.}
\end{figure}

\section*{Conclusions}
\paragraph{} In this study, we addressed smart deterministic-assisted bi-objective optimizations of electrode performance in different energy-like or power-like applications conditions. Our method predicts the ideal set of manufacturing parameters to use to produce the electrode in light of the specified type of battery applications. The entire method with support from physics-based and ML models. We used previously created synthetic data that contained highly representative process parameters for the slurry, drying, and calendering steps as well as electrode properties that were extracted from the 3D microstructures created by the ARTISTIC physics-based manufacturing process simulator to perform the comparison of optimization problems.In order to accurately predict the performance descriptors E and P as a function of the amount of active material, the solid content in the slurry, and the compression degree, the aforementioned data allowed for the development of a deterministic learning algorithm that was trained and evaluated on the synthetic dataset. This allowed for the analysis of additional manufacturing conditions that were not present in the synthetic dataset. Regardless of the convergence conditions, this deterministic learning allows the development of many optimization loops with a generally short computing time. In the end, a Bayesian optimization approach is used to quickly minimize the various objective functions. Finally, we emphasized how $P$ separates the reduction in energy density into two branches. Based on this, electrode optimization for alternative C-rates, which corresponds to applications for energy- or power-oriented battery cells, have been added to the bi-objective optimization. We determined that the SC\,\% and CD\,\% should be lower (50.8\,\% and 12.8\,\%, respectively) for the ideal electrode (\textit{i.e.}, optimization of $E$) than for optimization instances of $P$ at low or very low C-rate.The overall workflow follows the interest to study numerous battery optimizations conditions which are practically important in industry, R\&D, and labs without the need to generate additional synthetic data from the physical modeling.Furthermore, according to our potential scenario based on the data at hand, the definition of a battery optimization problem only pertains to the distribution of weights and how to balance them in order to achieve the specified electrochemical performance.

\paragraph{} As a perspective, we aspire to shift our work to more manufacturing conditions and parameters such as particle size distribution, drying conditions (slow and fast drying rates). In addition to the research of NMC111, the general methodology may be applied to different material chemistries while being closely related to data analysis. The utilization of our complete computational physics-based modeling workflow is possible as long as the slurry and electrode can be handled as particle assemblies. Its ability to direct experimental users in fabricating high-performance battery cells is anticipated to be advantageous for battery manufacturing. The work presented here, which was analyzed for the fabrication of LIBs, can be applied to fabricate and optimize other electrochemical energy technologies (such as solid-state batteries, zinc-air batteries, and fuel cells), in order to improve their electrochemical performances. It is critical to specify additional electrode qualities that will need to be improved before using alternative technologies, such as lithium metal electrodes. Moreover, we are convinced that our approach can be applied to study other synthetic datasets using physics-based models focusing on aging mechanisms, such as lithium plating or SEI formation \cite{65}. Regarding the various batches of possible applications, we believe that our approach paves the way towards the development of software infrastructures enabling the gathering of significant amount of data, leading to autonomous Machine Learning frameworks. Under this scenario, this represents a major pillar to perform on-the-fly optimizations of battery manufacturing processes, with the potential to merge synthetic and experimental datasets to embed physical knowledge into new smart experimental data acquisition techniques. We think that this ambitious but practical strategy has the potential to be a turning point in battery research and development and speed up the energy transition.

\subsubsection*{Materials availability}
\paragraph{} This study did not generate new unique materials.

\subsubsection*{Data availability}
\paragraph{} Codes and data are available upon reasonable request. The dataset will be made also available through the computational portal in the \href{https///www.erc-artistic.eu/computational-portal}{ARTISTIC project website}.

\section*{Acknowledgement}
\paragraph{} A.A.F. and C.L. acknowledge the European Union's Horizon 2020 research and innovation program for the funding support through the European Research Council (grant agreement 772873, ARTISTIC project). M.D., E.A. and A.A.F. acknowledge the ALISTORE European Research Institute for funding support. A.A.F. acknowledges the Institut Universitaire de France for the support. We acknowledge Utkarsh Vijay at LRCS for the proofreading of the article and useful comments.

\section*{Supplementary materials}
\paragraph{} The Supporting Information (SI) provides the empirical distributions of the different electrode properties, the comparison between the different bi-objective optimizations in terms of electrochemical performances and manufacturing parameters as well as their representativity within the synthetic dataset. SI includes also the scatter plots for pairwise representations for the different electrode properties being inputs of the p2D model from COMSOL Multiphysics and manufacturing parameters, as well as a separate plot for the comparison of $E$ and $P$ as a function of manufacturing parameters. SI includes two tables for technical details about the p2D, and a table for the bounds used as prior information of the Bayesian optimizations. Finally, SI includes the scalar fitness function.

\section*{Author Contributions Statement}
\paragraph{} Conceptualization, M.D., and A.A.F.; Methodology, M.D., and A.A.F.; Investigation, M.D., and C.L.; Software, M.D., and C.L.; Formal Analysis, M.D.; Writing - Original Draft, M.D., and C.L.; Writing - Review \& Editing, M.D., C.L., V.K., E.A., and A.A.F.; Funding Acquisition, E.A., and A.A.F.; Resources, M.D., A.A.F.; Supervision, E.A., and A.A.F; Project Administration, A.A.F.

\section*{Competing Interests}
\paragraph{} The authors declare that they have no known competing interest or personal relationships that could have appeared to influence the work reported in this article.

%
%
%
%
%

\bibliographystyle{unsrt}  
\bibliography{ms}

\end{document}